\pdfoutput=1

\documentclass[11pt]{article}

\usepackage[final]{acl}
\usepackage{caption}
\usepackage{times}
\usepackage{latexsym}
\usepackage{float}
\usepackage{enumitem}
\usepackage{array}
\usepackage{tabularx}
\usepackage{caption}
\usepackage{booktabs}

\usepackage[T1]{fontenc}

\usepackage[utf8]{inputenc}

\usepackage{microtype}

\usepackage{inconsolata}

\usepackage{graphicx}

\usepackage{tabularx}
\usepackage{booktabs}
\usepackage{multirow}
\title{The Impact of Role Design in In-Context Learning for Large Language Models}

\author{Hamidreza Rouzegar \and Masoud Makrehchi \\[2ex] 
\textbf{Department of Electrical, Computer and Software Engineering} \\ 
\textbf{Ontario Tech University} \\ 
Oshawa, ON, Canada \\ 
\texttt{hamidreza.rouzegar@ontariotechu.net} \\ 
\texttt{masoud.makrehchi@ontariotechu.ca}\\ 
\\
}

\begin{document}
\maketitle
\begin{abstract}
In-context learning (ICL) enables Large Language Models (LLMs) to generate predictions based on prompts without additional fine-tuning. While prompt engineering has been widely studied, the impact of role design within prompts remains underexplored. This study examines the influence of role configurations in zero-shot and few-shot learning scenarios using GPT-3.5 and GPT-4o from OpenAI and Llama2-7b and Llama2-13b from Meta. We evaluate the models' performance across datasets, focusing on tasks like sentiment analysis, text classification, question answering, and math reasoning. Our findings suggest the potential of role-based prompt structuring to enhance LLM performance.


\end{abstract}

\section{Introduction}
In recent years, the field of Large Language Models (LLMs) has seen remarkable advancements. Models such as GPT-3 \citep{brown2020language} and Llama \citep{touvron2023Llama} have showcased impressive capabilities in various natural language tasks, including question answering \citep{Rouzegar2024Generative}, sentiment analysis \citep{zhang2023sentiment}, and text classification \citep{abburi2023generative}.

A new paradigm known as In-Context Learning (ICL) has emerged in natural language processing (NLP) \citep{dong2022survey}. In ICL, LLMs generate predictions based on provided prompts, which usually include a few training examples. This capability is particularly valuable as it allows models to adapt dynamically to diverse tasks without the computational cost of retraining. 

While previous research has extensively studied prompt optimization techniques, much of the focus has been on content selection and structure rather than the explicit assignment of conversational roles.\citep{juneja2024task} The underlying mechanisms through which role-based prompting affects model behavior remain underexplored. Role design involves structuring prompts with distinct roles such as system, user, and assistant. Understanding how these roles influence model performance can provide valuable insights for optimizing prompt engineering strategies.

In this paper, we examine the impact of role design in zero-shot and few-shot experiments using two prominent instruction-tuned LLMs \citep{zhang2023instruction}, namely GPT-3.5 and GPT-4o from OpenAI, as well as Llama2-7b and Llama2-13b from Meta. By selecting a diverse set of LLMs, our study provides a broad perspective on the effectiveness of role-based prompting. We systematically evaluate the models’ performance across a range of natural language tasks, including sentiment analysis, text classification, question answering, and mathematical reasoning. Furthermore, our introduction of Structural Accuracy as a secondary metric offers a new lens to assess model compliance with intended prompt structures, contributing a novel evaluation dimension to prompt engineering research.

This study examines the impact of role-based prompt design across various datasets and LLMs, including tasks that involve complex reasoning. Our findings offer insights into optimal prompt design strategies, potentially improving LLM performance across various natural language processing and reasoning tasks. We observed that incorporating pre-defined roles without altering the main prompts often led to better overall performance. For more complex reasoning tasks, we implemented additional steps in our prompt design to address the increased complexity. The source code, implementation details, and all datasets used in our experiments are available in a GitHub repository to ensure the reproducibility of our results.\footnote{\href{https://github.com/hrouzegar/Role_Based_In-Context-Learning}{GitHub Code}}

\section{Related Works}
Significant research has been conducted in the field of prompt engineering and in-context learning. ICL refers to the ability of LLMs to generate predictions based on their input without additional fine-tuning. \citep{dong2022survey} While some research has focused on training and fine-tuning to improve ICL \citep{gu2023pre,li2024mend,shi2023context}, it generally operates at the inference level of LLMs, relying on prompt engineering rather than model modification. \citep{bhattamishra2023understanding}

ICL typically involves providing few-shot examples, and several aspects can be considered in this process: selection, formatting, and instruction. Selection methods focus on choosing optimal examples from the training data for few-shot learning. \citep{dong2022survey} These can include techniques based on k-nearest neighbors (KNN) \citep{liu2021makes}, using mutual information \citep{sorensen2022information}, perplexity estimation \citep{gonen2022demystifying}, or reinforcement learning (RL) \citep{zhang2022active}, for example selection.

Instruction engineering is another crucial aspect of ICL. Instruction engineering in ICL has seen various approaches: Instruction Induction generates instructions from input-output pairs \citep{honovich2022instruction}, APE optimizes instructions by treating them as programs \citep{zhou2022large}, and SELF-INSTRUCT improves instruction-following by using self-generated examples \citep{wang2022self}. These methods aim to enhance LLMs' ability to understand and follow complex directions.

Formatting plays a significant role in both ICL and broader prompt engineering techniques. Methods like Chain-of-Thought (CoT) prompting \citep{wei2022chain} provide reasoning demonstrations, allowing models to develop reasoning abilities naturally when given sufficiently complex tasks. Variants such as complex CoT \citep{fu2022complexity} and Auto-CoT \citep{zhang2022automatic} further refine these demonstrations under different conditions.

The Self-Ask method \citep{press2022measuring} improves performance by decomposing complex questions into simpler sub-questions, which the model answers sequentially. The Memory-of-Thought (MoT) \citep{li2023mot} approach involves pre-thinking on an unlabeled dataset and storing high-confidence thoughts as external memory, which the model can recall during testing to aid reasoning.

Super In-Context Learning (SuperICL) \citep{xu2023small} combines large language models with locally fine-tuned smaller models to improve performance on supervised tasks. These smaller models act as plug-ins, providing specific task knowledge, while the larger models focus on general language understanding.

Iterative Context-Aware Prompter (iCAP) \citep{wang2022iteratively} explores an iterative prompting framework that progressively elicits relevant knowledge from LLMs for multi-step inference. Similarly, the LEAST-TO-MOST Prompting method breaks down complex problems into a series of simpler subproblems and solves them in sequence \citep{zhou2022least}. The Task-Agnostic Prefix Prompt (TAPP) \citep{ye2024investigating} is a fixed prompt anchored to the beginning of every input, regardless of the target task, which facilitates zero-shot generalization.

Lastly, while recent work has explored role-play prompting to enhance LLMs’ reasoning abilities \citep{kong2023better}, these studies have primarily focused on applying pre-defined personas in zero-shot settings without systematically comparing their impact across different learning paradigms. 

Previous research has largely evaluated zero-shot and few-shot prompting independently, often assuming that few-shot learning consistently outperforms zero-shot. However, studies have shown that few-shot prompting does not always yield better results, depending on the task and example selection. In contrast to prior work, our study not only examines the effect of role-based prompt structuring but also introduces multiple few-shot configurations beyond the standard approach of providing three examples. By comparing these variations systematically, we demonstrate that role design plays a critical role in optimizing few-shot learning performance, providing new insights into effective prompt engineering strategies.

\section{Methodology and Experimental Setup}
\subsection{Datasets}
Following \citep{min2022rethinking} and \citep{rouzegar2024enhancing}, our experiments were conducted using samples from a diverse range of datasets to evaluate the performance of different prompt designs across various natural language tasks. To explore more challenging reasoning tasks, we also added a dataset of math word problems. Our experiments included the following datasets:

\begin{itemize}
    \item commonsense\_qa: \citep{talmor-etal-2019-commonsenseqa} A dataset designed for common-sense question answering.
    \item ai2\_arc: \citep{sheng2020investigating} A benchmark dataset used for evaluating question-answering capabilities.
    \item wiki\_movie\_plots: \citep{abbrivia_topicmap_movie_plots} A dataset containing movie plots used for genre classification.
    \item IMDB\_reviews: \citep{npathi_imdb_reviews} A sentiment analysis dataset comprising movie reviews with corresponding sentiment labels.
    \item MATH Dataset: \citep{hendrycksmath2021} A dataset to test reasoning skills with mathematical word problem questions.
\end{itemize}

\subsection{Prompt Designs}
We utilized four LLMs for our experiments: GPT-3.5-turbo-0125 and GPT-4o from OpenAI, and Llama2-7b-chat and Llama2-13b-chat from Meta. These models were selected because of their robust performance and widespread applicability in various NLP tasks. \citep{brown2020language} \citep{touvron2023Llama}

To investigate the impact of role design, we experimented with five basic prompt configurations across all tasks. In role design, the "system" provides high-level instructions that guide the overall task and set the context for the interaction. This includes specifying the format, the rules for the responses, or any additional context necessary for the task. The "user" presents specific queries or prompts that the model needs to respond to, effectively driving the interaction and simulating real-world usage scenarios. The "assistant" is the model response to user queries, which should adhere to the system's instructions and accurately respond to user queries (Table \ref{tab:example_prompt}). Our prompt configurations include:

\begin{table*}[h]
\centering
\small
\setlength{\tabcolsep}{5pt}
\renewcommand{\arraystretch}{1.2}
\begin{tabular}{|p{0.18\linewidth}|p{0.75\linewidth}|}
\hline
\multicolumn{2}{|c|}{
\begin{minipage}{0.95\linewidth}
\centering
\textbf{Task Instruction:} \textit{"Determine the genre of the movie based on the provided plot. For the plot provided, classify its genre as a single word (without other marks or words like 'genre:'), either Comedy, Action, Drama, or Horror."}
\end{minipage}
} \\
\hline
\textbf{Prompt Type} & \textbf{Example} \\
\hline
\textbf{ZeroU} &  
\textbf{User:}  \textcolor{gray}{[Task Instruction]} \textcolor{blue}{Plot[i]}  
\newline \textbf{LLM Output:} \textit{"Action"} \\
\hline
\textbf{ZeroSU} &  
\textbf{System:} \textcolor{gray}{[Task Instruction]}  
\newline \textbf{User:} \textcolor{blue}{Plot[i]}  
\newline \textbf{LLM Output:} \textit{"Drama"} \\
\hline
\textbf{FewU} &  
\textbf{User:} \textcolor{gray}{[Task Instruction]}  
\newline \textit{"Example 1: Plot[1] - Genre[1]"}  
\newline \textit{"Example 2: Plot[2] - Genre[2]"}  
\newline \textit{"Example 3: Plot[3] - Genre[3]"}  
\newline \textcolor{blue} {Plot[i]} 
\newline \textbf{LLM Output:} \textit{"Comedy"} \\
\hline
\textbf{FewSU} &  
\textbf{System:} \textcolor{gray}{[Task Instruction]}  
\newline \textbf{User:} \textit{"Example 1: Plot[1] - Genre[1]"}  
\newline \textit{"Example 2: Plot[2] - Genre[2]"}  
\newline \textit{"Example 3: Plot[3] - Genre[3]"}  
\newline \textcolor{blue} {Plot[i]}   
\newline \textbf{LLM Output:} \textit{"Horror"} \\
\hline
\textbf{FewSUA} &  
\textbf{System:} \textcolor{gray}{[Task Instruction]}  
\newline \textbf{User:} \textit{Plot[1]}  
\newline \textbf{Assistant:} \textit{Genre[1]}  
\newline \textbf{User:} \textit{Plot[2]}  
\newline \textbf{Assistant:} \textit{Genre[2]}  
\newline \textbf{User:} \textit{Plot[3]}  
\newline \textbf{Assistant:} \textit{Genre[3]}  
\newline \textbf{User:} \textcolor{blue} {Plot[i]}  
\newline \textbf{LLM Output:} \textit{"Action"} \\
\hline
\end{tabular}
\caption{Examples of different prompt designs used in this study. Each design modifies role assignment or omits roles while maintaining the same classification task.}
\label{tab:example_prompt}
\end{table*}

\begin{itemize}
\item Zero-shot User Prompt (ZeroU): The entire prompt is provided as a user instruction without any role distinctions.
\item Zero-shot System and User Prompt (ZeroSU): The initial part of the prompt is provided as a system instruction, followed by the user input.
\item Few-shot User Prompt (FewU): Multiple examples of questions and answers are provided within the user prompt.
\item Few-shot System and User Prompt (FewSU): Examples are provided with clear distinctions between system and user roles.
\item Few-shot System, User, and Assistant Prompt (FewSUA): Includes examples with system instructions, user inputs, and assistant responses.
\end{itemize} 

For the more complex math reasoning tasks, we extended our approach by implementing four sets of these five designs, each set incorporating additional prompt engineering steps:
\begin{enumerate}

    \item Basic Math: Applied our standard five configurations, tailored for math questions.
    \item Specialized Math: Refined the system instructions in each configuration to explicitly request solving the math question and providing the answer as a letter.
    \item Explanation Request: Added a request for the model to explain its rationale after providing the answer in each of the five configurations.
    \item Reasoning-First: Modified the order of operations in each configuration, asking the model to explain its reasoning process before providing the final answer.

\end{enumerate}

This approach resulted in a total of 20 different prompt designs for the math reasoning tasks, five basic configurations multiplied by four refinement steps. By maintaining the five role-based structures across each refinement step, we were able to examine both the effects of role-based prompting and the impact of task-specific instruction refinements on model performance.

\section{Experimental Setup}

\subsection{Data Preparation and Model Configuration}
For each dataset, we selected a balanced subset to ensure a comprehensive evaluation. For classification tasks (such as genre classification and sentiment analysis), we ensured a balanced distribution of categories and labels. For the question-answering tasks and math word problems, we maintained an equal distribution of answers across the multiple-choice options. For the few-shot scenario, we randomly selected three samples from each dataset, ensuring that these examples were not included in the test set. 

All GPT and Llama models were configured with a temperature of 0.0. This setting minimizes the randomness in the model's output, ensuring more deterministic responses. Other parameters, such as maximum sequence length and batch size, were optimized for each task to ensure optimal performance.

\subsection{Prompt Construction}
As demonstrated in Figure 1, we maintained a consistent format for prompt design in all five data sets, with minor configurations for each specific task. The core structure of each prompt consisted of two main parts:

\begin{enumerate}

    \item High-level instruction: This part provided the general task description and output format. For example, for the movie genre classification task, it read: "Determine the genre of the movie based on the provided plot. For the plot provided, classify its genre as a single word (without other marks or words like 'genre:'), either "comedy", "action", "drama", or "horror"." This instruction was modified for each dataset with simple word changes, such as "Determine the answer of the example based on the provided question" for question-answering tasks, with the output format adjusted accordingly (e.g., "a single small letter (without other marks or words like 'answer:'), either "a", "b", "c", "d", or "e".").
    \item Task-specific input: This part contained the actual content to be analyzed, such as the movie plot, review text, or question, always provided in the user prompt or at the end of the user prompt in few-shot experiments.
    
\end{enumerate}

\begin{table}[h]
\centering
\small
\setlength{\tabcolsep}{5pt}
\renewcommand{\arraystretch}{1.2}
\begin{tabular}{|p{0.95\linewidth}|}
\hline
\multicolumn{1}{|c|}{\textbf{Prompt Designs}} \\
\hline
\textbf{1. Zero-shot User (ZeroU):}
\begin{itemize} [topsep=3pt,partopsep=0pt,itemsep=0pt,parsep=0pt]
\setlength{\itemsep}{0pt}
\setlength{\parskip}{0pt}
\item All content in a single user prompt.
\end{itemize} \\
\hline
\textbf{2. Zero-shot System and User (ZeroSU):}
\begin{itemize}[topsep=3pt,partopsep=0pt,itemsep=0pt,parsep=0pt]
\setlength{\itemsep}{0pt}
\setlength{\parskip}{0pt}
\item High-level instruction as system prompt.
\item Task-specific input as user prompt.
\end{itemize} \\
\hline
\textbf{3. Few-shot User (FewU):}
\begin{itemize}[topsep=3pt,partopsep=0pt,itemsep=0pt,parsep=0pt]
\setlength{\itemsep}{0pt}
\setlength{\parskip}{0pt}
\item Examples and task-specific input in a single user prompt.
\end{itemize} \\
\hline
\textbf{4. Few-shot System and User (FewSU):}
\begin{itemize}[topsep=3pt,partopsep=0pt,itemsep=0pt,parsep=0pt]
\setlength{\itemsep}{0pt}
\setlength{\parskip}{0pt}
\item High-level instruction as system prompt.
\item Examples and task-specific input as user prompt.
\end{itemize} \\
\hline
\textbf{5. Few-shot System, User, and Assistant (FewSUA):}
\begin{itemize}[topsep=3pt,partopsep=0pt,itemsep=0pt,parsep=0pt]
\setlength{\itemsep}{0pt}
\setlength{\parskip}{0pt}
\item High-level instruction as system prompt.
\item Examples split between user and assistant prompts.
\item Followed by task-specific input.
\end{itemize} \\
\hline
\end{tabular}
\caption{Summary of Prompt Designs}
\label{tab:prompt_designs}
\end{table}

As outlined in Table \ref{tab:prompt_designs}, we implemented five distinct prompt designs for each dataset. These designs range from zero-shot to few-shot approaches, with varying degrees of role distinction. The zero-shot designs include ZeroU, which contains all content in a single user prompt, and ZeroSU, which separates high-level instructions into a system prompt. The few-shot designs incorporate examples, with FewU presenting these alongside task-specific input in a single user prompt. FewSU builds on this by distinguishing between system and user roles, while FewSUA further delineates roles by splitting examples between user and assistant prompts.

\begin{table*}[ht]
\centering
\captionsetup{justification=raggedright,singlelinecheck=false} 
{\footnotesize
\begin{tabularx}{\textwidth}{@{}l|l|>{\centering\arraybackslash}X|>{\centering\arraybackslash}X|>{\centering\arraybackslash}X|>{\centering\arraybackslash}X|>{\centering\arraybackslash}X|>{\centering\arraybackslash}X|>{\centering\arraybackslash}X|>{\centering\arraybackslash}X|>{\centering\arraybackslash}X|>{\centering\arraybackslash}X@{}}
\midrule[1.5pt]
\multicolumn{1}{c|}{\textbf{Dataset}} & \multicolumn{1}{c|}{\textbf{LLM}} & \multicolumn{2}{c}{\textbf{ZeroU}} & \multicolumn{2}{c}{\textbf{ZeroSU}} & \multicolumn{2}{c}{\textbf{FewU}} & \multicolumn{2}{c}{\textbf{FewSU}} & \multicolumn{2}{c}{\textbf{FewSUA}} \\ \cmidrule(l){3-12} 
\multicolumn{1}{c|}{} & \multicolumn{1}{c|}{} & Str. Acc. & F1 Score & Str. Acc. & F1 Score & Str. Acc. & F1 Score & Str. Acc. & F1 Score & Str. Acc. & F1 Score \\ \midrule[1.5pt]
\multirow{4}{*}{\textbf{commonsense\_qa}} & GPT-3.5 & 24 & 68 & 60 & 68 & 20 & 69 & 55 & 68 & 100 & \textbf{73} \\ \cmidrule{2-12} 
 & GPT-4o & 100 & 77 & 100 & 80 & 99 & 79 & 100 & 82 & 100 & \textbf{83} \\ \cmidrule{2-12} 
 & Llama2-7b & 0 & \textbf{19} & 0 & \textbf{19} & 0 & 18 & 0 & \textbf{19} & 67 & 9 \\ \cmidrule{2-12} 
 & Llama2-13b & 0 & 33 & 0 & \textbf{36} & 0 & \textbf{36} & 0 & \textbf{36} & 99 & 28 \\ \midrule[1.5pt]
\multirow{4}{*}{\textbf{ai2\_arc}} & GPT-3.5 & 73 & 76 & 39 & 80 & 60 & 80 & 87 & 78 & 99 & \textbf{85} \\ \cmidrule{2-12} 
 & GPT-4o & 99 & 96 & 100 & 96 & 100 & 95 & 100 & 96 & 100 & \textbf{97} \\ \cmidrule{2-12} 
 & Llama2-7b & 0 & 36 & 0 & 39 & 0 & 26 & 0 & 25 & 9 & \textbf{40} \\ \cmidrule{2-12} 
 & Llama2-13b & 0 & \textbf{50} & 0 & 52 & 0 & 39 & 0 & 37 & 77 & 48 \\ \midrule[1.5pt]
\multirow{4}{*}{\textbf{wiki\_movie\_plots}} & GPT-3.5 & 99 & 76 & 99 & \textbf{79} & 99 & 77 & 99 & 76 & 99 & 77 \\ \cmidrule{2-12} 
 & GPT-4o & 100 & 80 & 100 & 81 & 100 & 81 & 100 & 82 & 100 & \textbf{84} \\ \cmidrule{2-12} 
 & Llama2-7b & 0 & \textbf{75} & 0 & \textbf{75} & 0 & 68 & 0 & 70 & 25 & 74 \\ \cmidrule{2-12} 
 & Llama2-13b & 0 & 73 & 0 & 75 & 0 & 72 & 0 & 75 & 64 & \textbf{85} \\ \midrule[1.5pt]
\multirow{4}{*}{\textbf{IMDB\_reviews}} & GPT-3.5 & 100 & \textbf{94} & 100 & 93 & 100 & 93 & 100 & 67 & 100 & \textbf{94} \\ \cmidrule{2-12} 
 & GPT-4o & 100 & 95 & 100 & 95 & 100 & 96 & 100 & 92 & 100 & \textbf{97} \\ \cmidrule{2-12} 
 & Llama2-7b & 1 & \textbf{87} & 1 & 87 & 0 & 62 & 0 & 60 & 18 & 85 \\ \cmidrule{2-12} 
 & Llama2-13b & 3 & 91 & 3 & 90 & 0 & 64 & 0 & 67 & 82 & \textbf{93} \\ \midrule[1.5pt]
\end{tabularx}
}
\caption{\label{tab:LLM_results}
Performance results of various language models using different prompt designs on various datasets. The table includes Structural Accuracy (Str. Acc.) and F1 scores for each method: ZeroU (Zero-shot User-only), ZeroSU (Zero-shot System and User), FewU (Few-shot User-only), FewSU (Few-shot System and User), and FewSUA (Few-shot System, User, and Assistant).}
\end{table*}

\subsection{Adaptive Prompt Design}
In addition to our standard prompt designs, we implemented three more steps of prompt engineering for the MATH dataset, which represents a more complex reasoning task compared to the other question-answering datasets. The progression of these steps was as follows:

\begin{enumerate}

    \item Basic Math Prompt: This step used a prompt similar to our other datasets, with minor adjustments for math questions. The system instruction was: "Determine the answer of the example based on the provided question. For the question provided, classify its answer as a single small letter (without other marks or words like 'answer:'), either "a", "b", "c", "d", or "e"."
    \item Specialized Math Prompt: We refined the system instructions to more explicitly match the math problem-solving process. The new instruction was: "Solve the following math question and determine the correct answer. Provide the answer as a single small letter (without any additional marks or words like 'answer:'), choosing from: "a", "b", "c", "d", or "e"." This change aimed to better align the prompt with the mathematical nature of the task.
    \item Explanation Request: In this step, we added a request for the model to explain its rationale after providing the answer. The instruction was extended to: "After providing the letter answer, explain your rationale and how you arrived at the solution." In the few-shot examples for this configuration, we used the format of an answer followed by an explanation.
    \item Reasoning-First Approach: In the final configuration, we modified the order of operations, asking the model to explain its reasoning process before providing the final answer. The instruction was changed to: "Solve the following math question and determine the correct answer. Explain your rationale and how you came to the solution. After explaining how you reached the answer, provide the answer as a single small letter (without any additional marks), choosing from: "a", "b", "c", "d", or "e"."
    
\end{enumerate}

By implementing these progressive refinements in prompt design for math reasoning tasks, we created a spectrum of instruction complexities. This ranged from basic answer selection to detailed reasoning processes. This approach enabled us to assess how different cognitive demands embedded within the prompts influence the model performance in complex problem-solving scenarios. In addition, it provided information on how models adapt to increasingly sophisticated instruction sets while maintaining consistent role-based structures.

\subsection{Evaluation Metrics} 
The primary evaluation metric was the F1 score, which checks if the output of the prompt matches the class for sentiment analysis and movie genres or the correct answer on QA tasks. The second metric is structural accuracy, which measures how closely the output matches the requested structure in the prompt. For example, the movie genre should be a single word among the specified genres, the sentiment should be a single word, either "positive or "negative," and the QA answers should be a single capital letter among the choices (A, B, C, or D).

We examined structural accuracy regardless of the correctness of the answer. For instance, if the output was a single letter (A to D) for ai2\_arc, it was considered structured. However, outputs like "the answer to this question is D: state park" or simply "D: state park" were considered unstructured because they did not match the desired format. Additionally, for measuring the F1 score, we used post-processing to extract the label from the output and determine if it matched the true label of the dataset.

\begin{table*}[ht]
\centering
\captionsetup{justification=raggedright,singlelinecheck=false} 
{\footnotesize
\begin{tabularx}{\textwidth}{@{}l|l|>{\centering\arraybackslash}X|>{\centering\arraybackslash}X|>{\centering\arraybackslash}X|>{\centering\arraybackslash}X|>{\centering\arraybackslash}X|>{\centering\arraybackslash}X|>{\centering\arraybackslash}X|>{\centering\arraybackslash}X|>{\centering\arraybackslash}X|>{\centering\arraybackslash}X@{}}
\midrule[1.5pt]
\multicolumn{1}{c|}{\textbf{Prompt Design}} & \multicolumn{1}{c|}{\textbf{LLM}} & \multicolumn{2}{c}{\textbf{ZeroU}} & \multicolumn{2}{c}{\textbf{ZeroSU}} & \multicolumn{2}{c}{\textbf{FewU}} & \multicolumn{2}{c}{\textbf{FewSU}} & \multicolumn{2}{c}{\textbf{FewSUA}} \\ \cmidrule(l){3-12} 
\multicolumn{1}{c|}{} & \multicolumn{1}{c|}{} & Str. Acc. & F1 Score & Str. Acc. & F1 Score & Str. Acc. & F1 Score & Str. Acc. & F1 Score & Str. Acc. & F1 Score \\ \midrule[1.5pt]
\multirow{2}{*}{\textbf{Basic Math}} & GPT-3.5 & 0.03 & 0.24 & 0.03 & \textbf{0.38} & 0.62 & 0.24 & 0.80 & 0.24 & 1.00 & 0.26 \\ \cmidrule{2-12} 
 & GPT-4o & 1.00 & 0.33 & 1.00 & 0.35 & 1.00 & 0.36 & 1.00 & 0.37 & 1.00 & \textbf{0.40} \\ \midrule[1.5pt]
\multirow{2}{*}{\textbf{Specialized Math}} & GPT-3.5 & 0.27 & 0.31 & 0.00 & \textbf{0.53} & 0.96 & 0.23 & 0.25 & 0.29 & 1.00 & 0.24 \\ \cmidrule{2-12} 
 & GPT-4o & 0.74 & 0.39 & 0.63 & \textbf{0.42} & 0.90 & 0.35 & 0.99 & 0.35 & 1.00 & 0.41 \\ \midrule[1.5pt]
\multirow{2}{*}{\textbf{Explanation Request}} & GPT-3.5 & 0.00 & 0.44 & 0.00 & \textbf{0.47} & 0.63 & 0.25 & 0.02 & \textbf{0.47} & 0.99 & 0.25 \\ \cmidrule{2-12} 
 & GPT-4o & 0.29 & \textbf{0.49} & 0.30 & \textbf{0.49} & 1.00 & 0.31 & 1.00 & 0.32 & 0.99 & 0.40 \\ \midrule[1.5pt]
\multirow{2}{*}{\textbf{Reasoning-First}} & GPT-3.5 & N/A & 0.47 & N/A & 0.51 & N/A & 0.55 & N/A & 0.50 & N/A & \textbf{0.58} \\ \cmidrule{2-12} 
 & GPT-4o & N/A & 0.45 & N/A & 0.46 & N/A & 0.49 & N/A & 0.48 & N/A & \textbf{0.53} \\ \midrule[1.5pt]
\end{tabularx}
}
\caption{\label{tab:MATH_results1}
Performance results of GPT-3.5 and GPT-4o models using different prompt designs on the MATH dataset. The table includes Structural Accuracy (Str. Acc.) and F1 scores for each method: ZeroU (Zero-shot User-only), ZeroSU (Zero-shot System and User), FewU (Few-shot User-only), FewSU (Few-shot System and User), and FewSUA (Few-shot System, User, and Assistant). The prompt designs progress from Basic Math to Reasoning-First, reflecting increasing complexity in instruction structure.}
\end{table*}

\section{Results}
The results of our experiments are summarized in Tables \ref{tab:LLM_results} and \ref{tab:MATH_results1}, where we present the F1 scores and structural accuracy (Str. Acc.) for various prompt designs across five datasets, including the more complex MATH dataset. The experiments reveal several key insights regarding the impact of role design and prompt engineering in LLMs.

\subsection{General NLP Task Performance}

For the general NLP tasks (Table \ref{tab:LLM_results}), in most cases, the few-shot system, user, and assistant prompt (FewSUA) configuration exhibited a superior F1 score across the experiments, indicating that incorporating clear role distinctions and examples significantly enhances model performance.

For Llama models, the use of few-shot user (FewU) and few-shot system and user (FewSU) commands led to a noticeable decrease in F1 scores. This suggests that embedding few-shot examples within user prompts may cause these models to produce more incorrect answers and hallucinations.

The structural accuracy of GPT models was generally high, with both models adhering to the desired output structure in most cases. However, the Llama models frequently generated more than one word, failing to follow the instructions, which is understandable given their smaller size. A standout finding is that the FewSUA prompt configuration helped the Llama models generate outputs in the desired structure, making it potentially useful for applications like chatbots where maintaining a specific structure is crucial. In addition, Llama2-13b outperformed Llama2-7b in structural adherence.

Furthermore, in tasks such as movie genre classification and sentiment analysis, the Llama2-13b model showed results comparable to GPT models, demonstrating its capability in certain NLP tasks despite its smaller size. However, in question-answering tasks, the performance of the GPT models was significantly better and not comparable to that of the Llama models.

The results from these general NLP tasks highlighted the effectiveness of role-based prompting, particularly the FewSUA configuration. However, they also revealed some limitations, especially for smaller models like Llama. These findings prompted us to explore more refined prompt engineering techniques for more complex reasoning tasks, specifically in the domain of mathematical problem-solving. The following subsection details our approach and findings in this more challenging domain.

\subsection{Mathematical Reasoning Task Performance}

The results of our experiments on the MATH dataset, as presented in Table \ref{tab:MATH_results1}, reveal intriguing and sometimes unexpected patterns in the performance of GPT-3.5 and GPT-4o across different prompt designs.

In the Basic Math prompt, we observed an unusual trend with the superiority of the ZeroSU configuration, particularly for GPT-3.5 (F1 score of 0.38). The overall performance of other approaches was not as strong as anticipated. This led us to refine our prompt by making minor adjustments to better suit mathematical questions, resulting in the Specialized Math prompt.

The Specialized Math prompt showed a similar trend, with ZeroSU maintaining its edge (F1 score of 0.53 for GPT-3.5). Interestingly, we noticed that configurations with lower structural accuracy often yielded better performance in providing correct answers. Upon analyzing the model responses, we found that better-performing answers often included explanations and reasoning rather than just a single-letter response.

This observation prompted us to test a new hypothesis by incorporating reasoning and rationale into our prompts, leading to the Explanation Request design. In this configuration, we altered the few-shot examples to include explanations for each answer. The results showed that FewSU achieved better F1 scores, again with low structural accuracy. We observed that in some cases, the model made mistakes in attempting to maintain the structure, such as providing an initial answer, followed by an explanation, and then a different final answer (e.g., "a, [explanation], the answer is b").

To address these issues, we further evolved the prompt in the Reasoning-First approach, asking the model to explain its reasoning before providing the final answer. This design aimed to avoid the errors observed in the previous configuration. As a result, we did not evaluate structural accuracy for this prompt (marked as N/A in the table).

For GPT-4o, we observed similar trends, although the results across different configurations were more consistent, which aligns with its status as a more advanced LLM. 

Regarding the Llama models, we generally observed poor performance across the board, which is understandable given the complex nature of the mathematical reasoning tasks. However, an interesting exception was observed for Llama2-13b in the FewSUA configuration, which achieved a 31\% F1 score, significantly outperforming other configurations for this model.

These findings highlight the complex interplay between prompt design, model capabilities, and task complexity in mathematical reasoning. They also underscore the importance of allowing models to provide explanations and reasoning, which often leads to more accurate responses, even if at the cost of strict adherence to a prescribed output structure.

In general, our experiments reveal that the effectiveness of prompt design varies significantly between task types and model architectures. Although FewSUA configuration excelled in standard NLP tasks, mathematical reasoning required more nuanced approaches, often benefiting from explanatory prompts. Larger models consistently outperformed smaller ones, although Llama2-13b showed promise in certain NLP tasks. In particular, strict structural adherence did not always correlate with better performance, especially in complex reasoning tasks. These findings highlight the need for task-specific and model-specific prompt engineering strategies.

\section{Limitations and Future Works}

Our study has limitations that present opportunities for future research. We focused on a limited range of tasks and datasets. Future work should expand to include a broader array of NLP tasks, such as summarization and translation, to provide a more comprehensive understanding of the effectiveness of prompt design across different language processing challenges.

While we utilized pre-defined roles for instruction-tuned LLMs, future studies could explore creating new, task-specific roles to further optimize prompt designs. Additionally, investigating how these role-based strategies translate to non-instruction-tuned language models such as GPT-3 and Llama Base models could yield valuable insights into the broader applicability of this approach.

Due to resource limitations, we were unable to experiment with larger Llama models, which may show comparable results to the experimented GPT models in certain tasks. Future research with access to more computational resources could explore the performance of these larger models in role-based prompt designs, potentially revealing new insights into the scalability of this approach across different model sizes and architectures.

Lastly, future research could delve deeper into the relationship between structural accuracy and task performance, potentially leading to more optimized prompt engineering strategies. Exploring adaptive prompt designs that dynamically adjust based on the specific task and model could also prove fruitful.

\section*{Conclusion}
This study examines the impact of role-based prompt design and task-specific prompt engineering on Large Language Models (LLMs) across various NLP tasks and mathematical reasoning problems. Our experiments with GPT-3.5, GPT-4o, Llama2-7b, and Llama2-13b revealed that the few-shot system, user, and assistant (FewSUA) prompt configuration generally performed best in NLP tasks. However, in mathematical reasoning, we found that configurations allowing for explanations often outperformed those strictly adhering to structural accuracy.

Significant performance differences were observed between model sizes and architectures, with larger models generally performing better. Interestingly, in some simpler NLP tasks, Llama2-13b achieved results comparable to GPT models.

Our findings suggest that optimal prompt design varies not only between task types but also between models of different sizes and architectures. The success of the Reasoning-First approach in mathematical tasks indicates that encouraging models to articulate their thought processes can lead to more accurate outputs. These insights contribute to the growing understanding of effective prompt engineering for LLMs, offering valuable guidance for researchers and practitioners in the field of natural language processing.

\section*{Acknowledgements}
We gratefully acknowledge the support provided by the Natural Sciences and Engineering Research Council of Canada (NSERC) for their grant funding, which made this research possible. We also extend our sincere thanks to the Digital Research Alliance of Canada for providing the computational resources essential for conducting our experiments. Their support has been invaluable in enabling the extensive experiments conducted in this research.
\clearpage

\bibliography{anthology,custom}

\clearpage
\appendix

\label{sec:appendix}
\section{Appendix}
The appendix provides detailed descriptions and discussions of the examples of prompts and the outputs from different models, specifically for the wiki\_movie\_plots dataset. The prompt designs include ZeroU (Zero-shot User-only), ZeroSU (Zero-shot System and User), FewU (Few-shot User-only), FewSU (Few-shot System and User), and FewSUA (Few-shot System, User, and Assistant). Each design isolates the effect of role-based structuring on model performance. 

While the specific example in Figure \ref{Fig:prompts} is for the wiki\_movie\_plots dataset, similar prompts are used for other datasets with minor changes in the desired output, such as sentiment for IMDB reviews or answers for commonsense\_qa and ai2\_arc. By splitting the prompts into distinct roles, the study observes the impact on the models' ability to generate accurate and structured responses.

\begin{table*}[ht]
\centering
\captionsetup{justification=raggedright,singlelinecheck=false} 
{\footnotesize
\begin{tabularx}{\textwidth}{@{}l|l|>{\centering\arraybackslash}X|>{\centering\arraybackslash}X|>{\centering\arraybackslash}X|>{\centering\arraybackslash}X|>{\centering\arraybackslash}X|>{\centering\arraybackslash}X|>{\centering\arraybackslash}X|>{\centering\arraybackslash}X|>{\centering\arraybackslash}X|>{\centering\arraybackslash}X@{}}
\midrule[1.5pt]
\multicolumn{1}{c|}{\textbf{Prompt Design}} & \multicolumn{1}{c|}{\textbf{LLM}} & \multicolumn{2}{c}{\textbf{ZeroU}} & \multicolumn{2}{c}{\textbf{ZeroSU}} & \multicolumn{2}{c}{\textbf{FewU}} & \multicolumn{2}{c}{\textbf{FewSU}} & \multicolumn{2}{c}{\textbf{FewSUA}} \\ \cmidrule(l){3-12} 
\multicolumn{1}{c|}{} & \multicolumn{1}{c|}{} & Str. Acc. & F1 Score & Str. Acc. & F1 Score & Str. Acc. & F1 Score & Str. Acc. & F1 Score & Str. Acc. & F1 Score \\ \midrule[1.5pt]
\multirow{2}{*}{\textbf{Basic Math}} & Llama2-7 & 0.00 & 0.23 & 0.00 & 0.24 & 0.00 & 0.19 & 0.00 & 0.20 & 0.05 & 0.21 \\ \cmidrule{2-12} 
 & Llama2-13 & 0.00 & 0.23 & 0.00 & 0.23 & 0.00 & 0.20 & 0.00 & 0.20 & 0.71 & 0.21 \\ \midrule[1.5pt]
\multirow{2}{*}{\textbf{Specialized Math}} & Llama2-7 & 0.00 & 0.23 & 0.00 & 0.25 & 0.00 & 0.20 & 0.00 & 0.20 & 0.18 & 0.23 \\ \cmidrule{2-12} 
 & Llama2-13 & 0.00 & 0.22 & 0.00 & 0.23 & 0.00 & 0.20 & 0.00 & 0.20 & 0.68 & 0.21 \\ \midrule[1.5pt]
\multirow{2}{*}{\textbf{Explanation Request}} & Llama2-7 & 0.00 & 0.24 & 0.00 & 0.26 & 0.00 & 0.20 & 0.00 & 0.20 & 0.70 & 0.22 \\ \cmidrule{2-12} 
 & Llama2-13 & 0.00 & 0.19 & 0.00 & 0.20 & 0.00 & 0.20 & 0.00 & 0.20 & 0.93 & 0.21 \\ \midrule[1.5pt]
\multirow{2}{*}{\textbf{Reasoning-First}} & Llama2-7 & N/A & 0.16 & N/A & 0.22 & N/A & 0.20 & N/A & 0.20 & N/A & 0.23 \\ \cmidrule{2-12} 
 & Llama2-13 & N/A & 0.23 & N/A & 0.19 & N/A & 0.20 & N/A & 0.20 & N/A & \textbf{0.31} \\ \midrule[1.5pt]
\end{tabularx}
}
\caption{\label{tab:MATH_results2}
Performance results of Llama2-7 and Llama2-13 models using different prompt designs on the MATH dataset. The table includes Structural Accuracy (Str. Acc.) and F1 scores for each method: ZeroU (Zero-shot User-only), ZeroSU (Zero-shot System and User), FewU (Few-shot User-only), FewSU (Few-shot System and User), and FewSUA (Few-shot System, User, and Assistant). The prompt designs progress from Basic Math to Reasoning-First, reflecting increasing complexity in instruction structure.}
\end{table*}

Figures \ref{Fig:structured_responses}, \ref{Fig:gpt_responses}, and \ref{Fig:Llama_responses} present examples of structured and unstructured responses across different datasets, including sentiment analysis, genre classification, and question-answering tasks. Structured responses are critical for consistency in evaluation, requiring specific formats such as single words or letters. 

The post-processing technique mentioned in the text is essential for accurately extracting these labels from the model outputs, highlighted by the blue-colored text in the examples. However, there are instances where the extraction process makes mistakes, as shown in Figure \ref{Fig:gpt_responses}. For example, in the GPT-4o (commonsense\_qa) example, the post-processing incorrectly extracted 'A' as the answer, although the output indicated 'C'. This highlights the challenges in achieving precise output structures and the importance of improving post-processing techniques to avoid such errors.

The appendix emphasizes the importance of clear role distinctions in prompt design, particularly the FewSUA configuration, which consistently improves both F1 scores and structural adherence. The detailed examples and discussions provide valuable insights for developing effective prompt designs, highlighting the crucial role of post-processing in maintaining output accuracy.

\subsection{Resource Utilization}

For the experiments involving GPT models, we used the OpenAI API, with the total cost amounting to approximately 150 USD. For the Llama models, we utilized a single GPU with 64GB memory for the 7b model and two parallel GPUs, each with 64GB memory, for the 13b model. This setup ensured that we had sufficient computational resources to conduct the experiments efficiently and obtain reliable results.

\definecolor{customdarkgreen}{RGB}{0,100,0}
\definecolor{red}{RGB}{255,0,0}
\definecolor{blue}{RGB}{0,0,255}

\begin{figure*}[ht!]
\begin{tabularx}{\textwidth}{>{\raggedright\arraybackslash}X}
\toprule
\textcolor{red}{\textbf{ZeroU prompt:}} \\
\quad role: \textbf{"user"}, content: Determine the genre of the movie based on the provided plot: \\
\quad For the plot provided, classify its genre as a single word (without other marks or words like 'genre:'), either "comedy", "action", "drama", or "horror". \\
\quad \textcolor{blue}{Plot[i]} \\
\addlinespace
\textcolor{red}{\textbf{ZeroSU prompt:}} \\
\quad role: \textbf{"system"}, content: Determine the genre of the movie based on the provided plot. \\
\quad For the plot provided, classify its genre as a single word (without other marks or words like 'genre:'), either "comedy", "action", "drama", or "horror". \\
\quad role: \textbf{"user"}, content: \textcolor{blue}{Plot[i]} \\
\addlinespace
\textcolor{red}{\textbf{FewU prompt:}} \\
\quad role: \textbf{"user"}, content: Determine the genre of the movie based on the provided plot: \\
\quad For the plot provided, classify its genre as a single word (without other marks or words like 'genre:'), either "comedy", "action", "drama", or "horror". \\
\quad Examples: \\
\quad \quad \textcolor{blue}{Plot[1]} - \textcolor{blue}{Genre[1]} \\
\quad \quad \textcolor{blue}{Plot[2]} - \textcolor{blue}{Genre[2]} \\
\quad \quad \textcolor{blue}{Plot[3]} - \textcolor{blue}{Genre[3]} \\
\quad ... \\
\quad \textcolor{blue}{Plot[i]} \\
\addlinespace
\textcolor{red}{\textbf{FewSU prompt:}} \\
\quad role: \textbf{"system"}, content: Determine the genre of the movie based on the provided plot. \\
\quad For the plot provided, classify its genre as a single word (without other marks or words like 'genre:'), either "comedy", "action", "drama", or "horror". \\
\quad role: \textbf{"user"}, content: Examples: \\
\quad \quad \textcolor{blue}{Plot[1]} - \textcolor{blue}{Genre[1]} \\
\quad \quad \textcolor{blue}{Plot[2]} - \textcolor{blue}{Genre[2]} \\
\quad \quad \textcolor{blue}{Plot[3]} - \textcolor{blue}{Genre[3]} \\
\quad ... \\
\quad \textcolor{blue}{Plot[i]} \\
\addlinespace
\textcolor{red}{\textbf{FewSUA prompt:}} \\
\quad role: \textbf{"system"}, content: Determine the genre of the movie based on the provided plot. \\
\quad For the plot provided, classify its genre as a single word (without other marks or words like 'genre:'), either "comedy", "action", "drama", or "horror". \\
\quad role: \textbf{"user"}, content: \textcolor{blue}{Plot[1]} \\
\quad role: \textbf{"assistant"}, content: \textcolor{blue}{Genre[1]} \\
\quad role: \textbf{"user"}, content: \textcolor{blue}{Plot[2]} \\
\quad role: \textbf{"assistant"}, content: \textcolor{blue}{Genre[2]} \\
\quad role: \textbf{"user"}, content: \textcolor{blue}{Plot[3]} \\
\quad role: \textbf{"assistant"}, content: \textcolor{blue}{Genre[3]} \\
\quad role: \textbf{"user"}, content: \textcolor{blue}{Plot[i]} \\
\bottomrule
\end{tabularx}
\caption{\label{Fig:prompts}
This figure illustrates the various prompt designs utilized to feed all models. Each prompt design is defined as follows: ZeroU (Zero-shot User-only), ZeroSU (Zero-shot System and User), FewU (Few-shot User-only), FewSU (Few-shot System and User), and FewSUA (Few-shot System, User, and Assistant).}
\end{figure*}

\begin{figure*}[ht!]

\begin{tabularx}{\textwidth}{>{\raggedright\arraybackslash}X}
\toprule
\textcolor{red}{\textbf{Sentiment (IMDB reviews):}} \\
\quad \textcolor{blue}{positive} / \textcolor{blue}{negative} (single word) \\
\addlinespace
\textcolor{red}{\textbf{Genre (wiki\_movie\_plots):}} \\
\quad \textcolor{blue}{comedy}, \textcolor{blue}{drama}, \textcolor{blue}{action}, \textcolor{blue}{horror} (single word) \\
\addlinespace
\textcolor{red}{\textbf{QA (ai2\_arc):}} \\
\quad \textcolor{blue}{A} to \textcolor{blue}{E} (single letter) \\
\addlinespace
\textcolor{red}{\textbf{QA (commonsense\_qa):}} \\
\quad \textcolor{blue}{A} to \textcolor{blue}{D} (single letter) \\
\bottomrule
\end{tabularx}
\caption{\label{Fig:structured_responses} This figure illustrates the various structured responses for different datasets used to evaluate the models.}

\end{figure*}
\captionsetup{justification=raggedright,singlelinecheck=false}
\begin{figure*}[ht!]

\begin{tabularx}{\textwidth}{>{\raggedright\arraybackslash}X}
\toprule
\textcolor{red}{\textbf{GPT-4o (IMDB reviews):}} \\
\quad \textcolor{blue}{neutral} \\
\addlinespace
\textcolor{red}{\textbf{GPT-4o (commonsense\_qa):}} \\
\quad Greg's performance was mediocre. What might his habits be like? \textcolor{blue} {A}: distinguished B: unique C: peculiar D: exceptional E: outstanding \\
\quad The answer is: C \\
\addlinespace
\textcolor{red}{\textbf{GPT-3.5 (IMDB reviews):}} \\
\quad {science fiction} \\
\addlinespace
\textcolor{red}{\textbf{GPT-3.5 (commonsense\_qa):}} \\
\quad\textcolor{blue} {D}: to remove toxic material from the stomach before it is absorbed \\
\bottomrule
\end{tabularx}
\caption{\label{Fig:gpt_responses} This figure shows examples of unstructured responses from GPT models. The blue-colored text highlights the extracted labels in the examples.}

\end{figure*}

\begin{figure*}[ht!]

\begin{tabularx}{\textwidth}{>{\raggedright\arraybackslash}X}
\toprule
\textcolor{red}{\textbf{Llama2-7b (commonsense\_qa):}} \\
\quad Based on the question provided, the answer is: \\
\quad A: halving the distance between them \\
\quad Therefore, the answer is \textcolor{blue}{"A"}. \\
\addlinespace
\textcolor{red}{\textbf{Llama2-7b (IMDB reviews):}} \\
\quad The sentiment of the review is \textcolor{blue} {negative}. The reviewer expresses disappointment with the sequel compared to the original, citing a lack of fun and too much time spent on plot development. They also mention that Tim Thomerson's wisecracking is toned down, and the performances are adequate but the script lets them down. The action is described as routine and the plot is only mildly interesting, which suggests that the reviewer was not impressed with the film's action sequences or story. The reviewer also states that they need lots of silly laughs to stay entertained during a Trancers movie, but the film fails to deliver on this front. Overall, the review is negative and suggests that the reviewer did not enjoy the sequel as much as the original. \\
\addlinespace
\textcolor{red}{\textbf{Llama2-7b (wiki\_movie\_plots):}} \\
\quad Based on the provided plot, I would classify the movie as a \textcolor{blue} {comedy}. The plot involves a group of irresponsible and wealthy individuals who die in an accident and become ghosts, with the goal of freeing their friend from a regimented lifestyle in order to enter heaven. The movie appears to have a lighthearted and comedic tone, with elements of slapstick humor, witty dialogue, and farcical situations. The characters' antics and misadventures are likely to provide most of the comedic elements in the movie. \\
\addlinespace
\textcolor{red}{\textbf{Llama2-13b (wiki\_movie\_plots):}} \\
\quad Sure! Based on the plot provided, the genre of the movie would be: \\
\quad \textcolor{blue}{Comedy}. \\
\quad The plot involves a man trying to relive his youth and pursue a beautiful woman, with his friends, in a series of comedic and awkward situations. The tone of the plot is lighthearted and humorous, indicating that the movie would be a comedy. \\
\addlinespace
\textcolor{red}{\textbf{Llama2-13b (IMDB reviews):}} \\
\quad Based on the review, I would rate the movie 3 out of 10. Here's a breakdown of the review: \\
\quad * The reviewer can't knock the movie too hard because it's clear that the filmmakers were trying to make it bad or campy. \\
\quad * Some of the parts that were intended to be funny actually are, but only for their cheesiness. \\
\quad * The reviewer notes that the movie goes over the top near the end with a time warp and sexual innuendoes. \\
\quad * The reviewer is surprised that the movie got a PG rating, implying that it's not suitable for all audiences. \\
\quad * The reviewer concludes that it's a fun way to waste a couple of hours, but it's not on the same level as Star Wars or Trek. \\
\addlinespace
\textcolor{red}{\textbf{Llama2-13b (ai2\_arc):}} \\
\quad Sure! Based on the question provided, the answer is: \\
\quad \textcolor{blue}{D} : wildfire\\
\bottomrule
\end{tabularx}
\caption{\label{Fig:Llama_responses}
This figure shows examples of unstructured responses from Llama2 models. The blue-colored text highlights the extracted labels in the examples.}
\end{figure*}

\end{document}